\documentclass[runningheads]{llncs}
\usepackage{amsmath,bm}
\usepackage{graphicx}
\usepackage{color}
\usepackage{multirow}
\usepackage{subfigure}

\begin{document}
\title{Intestinal Parasites Classification Using Deep Belief Networks}
\titlerunning{Intestinal Parasites Classification Using Deep Belief Networks}
\author{Mateus Roder\inst{1}\orcidID{0000-0002-3112-5290} \and
Leandro A. Passos\inst{1}\orcidID{0000-0003-3529-3109} \and
Luiz Carlos Felix Ribeiro\inst{1}\orcidID{0000-0003-1265-0273} \and
Barbara Caroline Benato\inst{2}\orcidID{0000-0003-0806-3607} \and
Alexandre Xavier Falc\~{a}o\inst{2}\orcidID{0000-0002-2914-5380} \and
Jo\~{a}o Paulo Papa\inst{1}\orcidID{0000-0002-6494-7514}\\
%
%
\email{\{mateus.roder, leandro.passos, joao.papa\}@unesp.br}\\
\email{\{barbara.benato\}@students.ic.unicamp.br} \\
\email{\{afalcao\}@unicamp.br} \\
\institute{School of Sciences, S\~ao Paulo State University, Bauru, Brazil \and Institute of Computing, University of Campinas, Campinas, Brazil}}

\authorrunning{M. Roder et al.}
\maketitle              
\begin{abstract}
Currently, approximately $4$ billion people are infected by intestinal parasites worldwide. Diseases caused by such infections constitute a public health problem in most tropical countries, leading to physical and mental disorders, and even death to children and immunodeficient individuals. Although subjected to high error rates, human visual inspection is still in charge of the vast majority of clinical diagnoses. In the past years, some works addressed intelligent computer-aided intestinal parasites classification, but they usually suffer from misclassification due to similarities between parasites and fecal impurities. In this paper, we introduce Deep Belief Networks to the context of automatic intestinal parasites classification. Experiments conducted over three datasets composed of eggs, larvae, and protozoa provided promising results, even considering unbalanced classes and also fecal impurities.
\end{abstract}

\section{Introduction}
\label{s.introduction}

Estimates reveal that around $4$ billion people in the world are infected with some intestinal parasite~\cite{paho:2007}. The human intestinal parasitism is a public health problem, especially in tropical countries~\cite{who:2010}, in which such infections can lead children and immunodeficient adults to death. The detection and diagnosis of human intestinal parasitosis depend on the visual analysis of optical microscopy images obtained from fecal samples mostly. However, the manual analysis of those images is time-consuming and error-prone. In order to circumvent this problem, Suzuki et al.~\cite{Suzuki:2013a} proposed a fully automated enteroparasitosis diagnosis system via image analysis, which addressed the $15$ most common species of protozoa and helminths in Brazil. The proposed approach is composed of three main steps: (i) image segmentation, (ii) object delineation, and its further (iii) classification.

Previous works have also investigated protozoa and helminth parasites classification. Suzuki et al.~\cite{Suzuki:2013b}, for instance, introduced the Optimum Path Forest~\cite{PapaIJIST:09,PapaPR:12} classifier for such a task, with results that outperformed Support Vector Machines and Artificial Neural Networks. Later on, Peixinho et al.~\cite{Peixinho:2015} explored Convolutional Neural Networks (CNNs) in this context. Further, Peixinho et al.~\cite{peixinho2018delaunay} proposed generating synthetic samples to increase the number of images for under-represented classes by adding points onto a 2D projection space. Furthermore, Benato et al.~\cite{benato2018semi} investigated an approach to cope with the lack of supervised data by interactively propagating labels to reduce the user effort in data annotation. Finally, Castelo et al.~\cite{Castelo:2019} used bag of visual words to extract key points from superpixel-segmented images and further build a visual dictionary to automatic classify intestinal parasites.

Apart from the techniques mentioned earlier, Restricted Boltzmann Machines (RBMs)~\cite{Smolensky:86} obtained notorious attention due to their promising results in a wide variety of tasks such as data reconstruction~\cite{passosIJCNN:19}, exudate identification in retinal images~\cite{khojasteh2019exudate}, and collaborative filtering~\cite{salakhutdinov2007restricted}, to cite a few. Moreover, RBMs can be used as the building block for more complex and deep models such as Deep Belief Networks (DBNs)~\cite{Hinton:06} and Deep Boltzmann Machines (DBMs)~\cite{salakhutdinov2009deep}.

However, as far as we are concerned, no work has employed RBM-based models in the task of intestinal parasite classification to date. Therefore, the main contributions of this work are threefold: (i)~to propose an effective method for parasite classification using RBMs and DBNs; (ii)~to evaluate the ability of Restricted Boltzmann Machines ability for data augmentation; and (iii)~to foster the scientific literature concerning both RBM-based applications and intestinal parasites  identification.

The remainder of this paper is organized as follows: Section~\ref{s.theoretical} introduces the theoretical background concerning RBMs and DBNs, while Sections~\ref{s.methodology} and~\ref{s.results} present the methodology and the experimental results, respectively. Finally, Section~\ref{s.conclusion} states conclusions and future works.

\section{Theoretical Background}
\label{s.theoretical}

In this section, we provide a brief description of the main concepts regarding RBM and DBN formulations, as well as their discriminative variant to deal with classification problems.

\subsection{Restricted Boltzmann Machines}
\label{ss.rbm}

Restricted Boltzmann Machines stand for energy-based neural networks that can learn the probability distribution over a set of input vectors. Such models are named after the Boltzmann distribution, a measurement that uses the system's energy to obtain the probability of a given state. Energy-based models are inspired by physics since they assign a scalar energy value for each variable configuration, thus learning by adjusting their parameters to minimize the energy of the system. Moreover, they are modeled as a bipartite graph, i.e., there are no connections between units from the same layer. Such a technique assumes binary-valued nodes, although there are extensions to real- and even complex-valued inputs~\cite{srivastava2012,nakashika2017}. 

Given an initial configuration $(\bm{v},\bm{h})$, the energy of the system can be computed as follows:

\begin{equation}
 \label{e.energy}
 E(\bm{v},\bm{h}) = -\sum_{i=1}^mb_iv_i-\sum_{j=1}^nc_jh_j-\sum_{i=1}^m\sum_{j=1}^nW_{ij}v_ih_j,
\end{equation}
where $\bm{v}\in\Re^m$ and $\bm{h}\in\Re^n$ stand for the visible and hidden layers, respectively, and $\bm{b}\in\Re^m$ and $\bm{c}\in\Re^n$ denote their bias vectors. Additionally, $\bm{W}_{m\times n}$ corresponds to the weight matrix concerning the connections between layers $\bm{v}$ and $\bm{h}$.

The learning procedure aims at finding $\bm{W}$, $\bm{a}$, and $\bm{b}$ in such a way Equation~\ref{e.energy} is minimized. However, calculating the joint probability of the model is intractable since it requires computing every possible initial configuration. Moreover, one can estimate the conditional probabilities using alternated iterations over a Monte Carlo Markov Chain (MCMC) approach, where the probabilities of both input and hidden units can be computed as follows:

\begin{equation}
\label{e.ph}
p(h_j=1|\bm{v}) = \sigma\left(c_j + \sum_{i=1}^mW_{ij}\bm{v}_i\right),
\end{equation}
and
\begin{equation}
\label{e.pv}
p(v_i=1|\bm{h}) = \sigma\left(b_i + \sum_{j=1}^nW_{ij}\bm{h}_j\right),
\end{equation}
where $\sigma$ stands for the logistic-sigmoid function. Since the visible and hidden units are conditionally independent, one can train the network using the MCMC algorithm with Gibbs sampling through Contrastive Divergence (CD)~\cite{Hinton:02}. 

\subsection{Deep Belief Networks}
\label{ss.dbn}

\begin{figure}[!ht]
\centerline{\begin{tabular}{c}
\includegraphics[width=3.7cm]{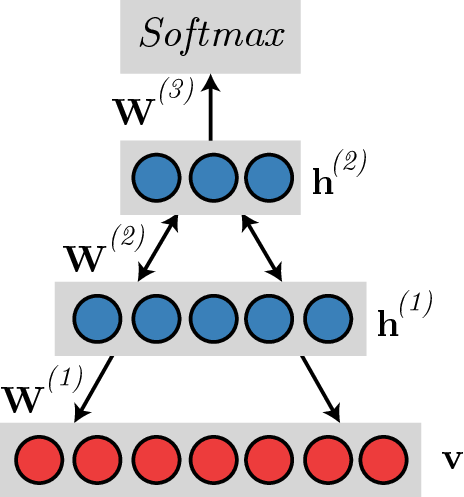} \\
\end{tabular}}
\caption{DBN architecture with two hidden layers for classification purposes.}
\label{f.dbn}
\end{figure}

Restricted Boltzmann Machines can also be employed to compose more complex models. They are commonly used as building blocks to generate the so-called Deep Belief Networks~\cite{Hinton:06}, which are composed of a visible and a set of $L$ hidden layers. In this model, each layer is connected to the next through a weight matrix $\textbf{W}^{(l)}$, $l \in [1,L]$. In short, DBNs consider each set of two subsequent layers as an RBM trained in a greedy fashion, where the hidden layer of the bottommost RBM feeds the next RBM's visible layer. For classification purposes, a Softmax layer is appended to the model. Afterwards, the model is fine-tuned using the backpropagation algorithm, as depicted in Figure~\ref{f.dbn}. Notice that $\textbf{h}^{(l)}$ stand for the $l$-th hidden layer.

\section{Methodology}
\label{s.methodology}

In this section, we introduce the dataset employed in this work, as well as the technical details concerning the experimental setup.

\subsection{Dataset}
\label{ss.datasets}

The experiments consider datasets from human intestinal parasites divided into three groups: (i) \textbf{Helminth eggs} (i.e., Eggs) with $12,691$ images, (ii) \textbf{Helminth larvae} (i.e., Larvae) with $1,598$ images, and (iii) \textbf{Protozoan cysts} (i.e., Protozoa) with $37,372$ images. Notice that all datasets contain fecal impurities, which is a diverse class that looks alike to some parasites. Each dataset comprises the following categories and their respective label in parenthesis:

\begin{sloppypar}
\begin{itemize}
	\item \textbf{Helminth eggs}: \emph{H.nana} (1), \emph{H.diminuta} (2), \emph{Ancilostomideo} (3), \emph{E.vermicularis} (4), \emph{A.lumbricoides} (5), \emph{T.trichiura} (6),  \emph{S.mansoni} (7), \emph{Taenia} (8), and impurities (9).
	\item \textbf{Helminth larvae}: larvae (1) and impurities (2); and
	\item \textbf{Protozoan cysts}: \emph{E.coli} (1), \emph{E.histolytica} (2), \emph{E.nana} (3), \emph{Giardia} (4), \emph{I.butschlii} (5), \emph{B.hominis} (6), and impurities (7).
\end{itemize}
\end{sloppypar}

These are the most common species of human intestinal parasites in Brazil, and they are also responsible for public health problems in most tropical countries~\cite{Suzuki:2013a}. Notice that all datasets are unbalanced with considerably more impurity samples. The objects of interest were first segmented from the background, converted to grayscale, and further resized to $50 \times 50$ pixels. Table~\ref{t.dist}(a) presents the distribution of samples per class.

\subsection{Data augmentation}
\label{ss.dataAugmentation}

In this paper, we proposed two different synthetic data generation approaches to overcome the class imbalance problem: (i) an Autoencoder (AE) and (ii) an additional RBM for image reconstruction purposes. In all cases, the models were trained with examples of the class to be oversampled only. Further, to allow a fair comparison, both the RBM and the AE contain similar architectures. Table~\ref{t.generator_hyperparameters} presents the hyperparameters employed while training the models for data augmentation.

\begin{table}[ht]
\renewcommand{\arraystretch}{1}
\centering
\begin{tabular}{llcc}
    \hline
        \textbf{Model}           &
        \textbf{Hyper-parameter} &
        \textbf{Search interval} &
        \textbf{Best value} \\
    \hline
        \multirow{4}{*}{AE}
        & $\eta$                  & $[10^{-5}, 10^{-2}]$     & $10^{-3}$  \\
        & $p_{drop}$              & $[0, 0.4]$               & $0.2$      \\
       & Hidden dim              & $\{250, 500, 2000\}$      & $500$      \\
        & Batch size              & $\{16, 32, 128\}$        & $32$       \\
    \hline
        \multirow{3}{*}{RBM}
        & $\eta$                  & $[10^{-5}, 10^{-2}]$     & $10^{-4}$  \\
        & Hidden dim              & $\{500, 2000\}$          & $500$      \\
        & Batch size              & $\{4, 8, 16\}$           & $8$        \\
\hline
        
\end{tabular}
\\~\\
\caption{Hyper-parameter setting up.}
\label{t.generator_hyperparameters}
\end{table}

Regarding the synthetic data generation, our policy is to oversample the minority classes in which the sum of total samples generated, for all classes, does not overpass approximately $50\%$ of the majority class (impurities). Table~\ref{t.dist}(b) presents the augmentation results. 

\begin{table}[!ht]
	\centering
	\resizebox{\textwidth}{!}{%
	\subfigure[Original]{
		\begin{tabular}{crrr}
			\hline
			\multirow{2}{*}{\textbf{Class}} &
			\multicolumn{3}{c}{\textbf{\# samples}}  \\
			&
			\multicolumn{1}{c}{Eggs} &
			\multicolumn{1}{c}{Larvae} &
			\multicolumn{1}{c}{Protozoa} \\
			\hline
			1    & 500   & 246    & 868   \\
			2    & 83    & 1,352  & 659   \\
			3    & 286   & --     & 1,783 \\
			4    & 103   & --     & 1,931 \\
			5    & 835   & --     & 3,297 \\
			6    & 435   & --     & 309   \\
			7    & 254   & --     & 28,525\\
			8    & 379   & --     & --    \\
			9    & 9,816 & --     & --    \\
			\hline
			\textbf{Total}  &
			\textbf{12,691} &
			\textbf{1,598}  &
			\textbf{37,372} \\
			\hline
	\end{tabular}}
	\centering
	\subfigure[Augmented]{
		\begin{tabular}{crrr}
			\hline
			\multirow{2}{*}{\textbf{Class}} &
			\multicolumn{3}{c}{\textbf{\# samples}}  \\
			&
			\multicolumn{1}{c}{Eggs} &
			\multicolumn{1}{c}{Larvae} &
			\multicolumn{1}{c}{Protozoa} \\
			\hline
			1    & 1,000 (500)  & 738 (492)   & 868   \\
			2    & 415 (332)   & 1,352  & 1,977 (1,318)  \\
			3    & 572 (286)  & --     & 1,783   \\
			4    & 412 (309)   & --     & 1,931   \\
			5    & 835   & --     & 3,297   \\
			6    & 870 (435)   & --     & 1,236 (927)   \\
			7    & 2,508 (2,254)   & --   & 28,525 \\
			8    & 379   & --     & --      \\
			9    & 9,816 & --     & --      \\
			\hline
			\textbf{Total}  &
			\textbf{14,807 (2,116)} &
			\textbf{2,090 (492)} &
			\textbf{39,619 (2,245)} \\
			\hline
	\end{tabular}}}
	\caption{Class frequency regarding the (a) original and (b) augmented datasets. The values in parenthesis stand for the number of samples generated artificially.}
	\label{t.dist}
\end{table}
\subsection{Experimental Setup}
\label{ss.experimental_setup}

Three different models were considered in this paper: one RBM with $500$ hidden neurons and two DBNs, i.e., the first with two hidden layers (DBN-2) containing $500$ neurons each, and the other comprising three hidden layers (DBN-3) with $2,000$ neurons in the first two levels and $500$ neurons in the uppermost layer\footnote{In case of acceptance, we shall provide the link to the source-code.}. All models were trained for $100$ epochs considering each RBM stack with a learning rate $\eta=10^{-5}$ and mini-batches of $64$ samples. Further, the networks were fine-tuned for an additional $100$ epochs with mini-batches of size $128$.

\subsection{Evaluation procedure}
\label{ss.evaluation}

Since we have unbalanced datasets, the standard accuracy (ACC) may not be suitable to evaluate the proposed models since it favors classifiers biased towards the most common classes. To address such an issue, we considered the Balanced Accuracy score (BAC)~\cite{Brodersen:2010} implemented in \texttt{sklearn}\footnote{Available at \url{https://scikit-learn.org}.}. Additionally, the Cohen's kappa coefficient~\cite{fleiss1973equivalence} is employed to assess the degree of agreement between the classifier and the ground truth labels. Such a value lies in the interval $[-1, 1]$, where the lower and upper boundaries represent a complete disagreement and an agreement, respectively. Finally, we employed the Wilcoxon signed-rank test~\cite{Wilcoxon:45} with significance of $5\%$ to evaluate the statistical similarity among the best results.




\color{black}

\section{Experimental Results}
\label{s.results}

In this section, we present the experimental results concerning automatic human parasites classification.

\subsection{Classification results}
\label{ss.ClassificationResults}

Table~\ref{t.larvae_summary} presents the mean results, concerning the standard accuracy, the balanced accuracy, and the Kappa value with respect to the Larvae dataset. Results are presented over the RBM, DBN-2, and DBN-3 techniques using three distinct configurations, i.e., the original dataset and its augmented versions using RBM (Aug-RBM) and AE (Aug-AE). Moreover, the best ones regarding Wilcoxon test are in bold.

The results confirm the robustness of the proposed approaches since all models with RBM Augmentor achieved more than $94\%$ of BAC. One can highlight the DBN-2 results using the Aug-RBM with $95\%$ and $0.901$ of mean accuracy and Kappa values, respectively. Such results provide good shreds of evidence towards the relevance of data augmentation with respect to the baseline, once Aug-RBM supported an improvement of around $5.6\%$ concerning the standard accuracy, $17.3\%$ regarding BAC, and $38\%$ considering the Kappa value. Although Aug-AE provided some improvements, RBM figures as the most accurate approach for such a task.

\begin{table*}[!htb]
\renewcommand{\arraystretch}{1.85}
\centering
\resizebox{\textwidth}{!}{
\begin{tabular}{llllllllllllllll}
\hline
& \multicolumn{3}{c}{\textbf{RBM}} & \multicolumn{3}{c}{\textbf{DBN-2}} & \multicolumn{3}{c}{\textbf{DBN-3}} \\ \hline
                          				& Aug-RBM          & Aug-AE       & Baseline                  & Aug-RBM           & Aug-AE         & Baseline       & Aug-RBM   &    Aug-AE                &       Baseline     \\ \hline    
                          				       
\multicolumn{1}{c}{\textbf{ACC}}   & 94.03$\pm$0.30 & 77.03$\pm$1.85 & 90.14$\pm$0.14   & \textbf{95.05$\pm$0.34} & 90.66$\pm$0.87  & 90.53$\pm$0.23  & 94.85$\pm$0.33 & 92.15$\pm$0.65  & 89.61$\pm$1.26  \\

\multicolumn{1}{c}{\textbf{BAC}}   & 94.07$\pm$0.28 & 69.71$\pm$2.95 & 80.19$\pm$0.38   & \textbf{95.09$\pm$0.33} & 90.63$\pm$0.75  & 81.24$\pm$0.41  & \textbf{94.87$\pm$0.34} & 91.40$\pm$0.79 & 80.99$\pm$2.29 \\

\multicolumn{1}{c}{\textbf{Kappa}} & 0.880$\pm$0.005 & 0.445$\pm$0.053 & 0.637$\pm$0.006  & \textbf{0.901$\pm$0.007} & 0.804$\pm$0.018  & 0.653$\pm$0.007 & \textbf{0.897$\pm$0.007} & 0.832$\pm$0.014 & 0.630$\pm$0.041 \\ \hline

\end{tabular}}
\\~\\
\caption{Effectiveness over Larvae dataset using the proposed approaches.}
\label{t.larvae_summary}
\end{table*}

Table~\ref{t.eggs_summary} presents the results regarding the Eggs dataset. In this scenario, DBN-3 obtained the best results concerning the ACC and Kappa values, while the standard RBM performed better over the BAC measure. This behavior is surprising since both Kappa and BAC were proposed to cope with unbalanced data evaluation, thus expecting to behave similarly to the other models.


\begin{table*}[!htb]
\renewcommand{\arraystretch}{1.85}
\centering
\resizebox{\textwidth}{!}{
\begin{tabular}{llllllllllllllll}
\hline
& \multicolumn{3}{c}{\textbf{RBM}} & \multicolumn{3}{c}{\textbf{DBN-2}} & \multicolumn{3}{c}{\textbf{DBN-3}} \\ \hline
                          				& Aug-RBM          & Aug-AE       & Baseline                  & Aug-RBM           & Aug-AE         & Baseline       & Aug-RBM   &    Aug-AE                &       Baseline     \\ \hline 
\multicolumn{1}{c}{\textbf{ACC}}   & 93.54$\pm$0.37 & 84.25$\pm$1.13  & 90.30$\pm$0.052  & 94.03$\pm$0.19 & 92.13$\pm$0.99 & 91.91$\pm$0.45 & \textbf{94.41$\pm$0.32} & 94.01$\pm$0.19 & 93.08$\pm$0.31  \\

\multicolumn{1}{c}{\textbf{BAC}}   & \textbf{92.09$\pm$0.68} & 67.15$\pm$2.54  & 79.94$\pm$0.55  & 90.98$\pm$0.77 & 88.36$\pm$1.77 & 78.34$\pm$1.33  & 91.06$\pm$0.62 & 90.39$\pm$0.30 & 78.67$\pm$1.75 \\

\multicolumn{1}{c}{\textbf{Kappa}} & 0.884$\pm$0.006 & 0.685$\pm$0.025 & 0.769$\pm$0.009  & 0.891$\pm$0.004 & 0.857$\pm$0.015  & 0.794$\pm$0.009 & \textbf{0.897$\pm$0.006}  & 0.890$\pm$0.003  & 0.820$\pm$0.009 \\ \hline

\end{tabular}}
\\~\\
\caption{Effectiveness over Eggs dataset using the proposed approaches.}
\label{t.eggs_summary}
\end{table*}

The behavior observed in the Protozoa dataset, presented in Table~\ref{t.proto_summary}, highlights an interesting scenario. One of the best ACC ($87.51\%$) and Kappa ($0.736$) results were achieved with the simplest model, i.e., an RBM using Aug-RBM. Such behavior points out that, for such a dataset, we can compress the input data into a small latent space, thus extracting useful and representative features with only $500$ units, while the performance is still remarkable even with unbalanced classes. Moreover, concerning BAC values, one can observe that DBN-2 and DBN-3 with data augmentation by Restricted Boltzmann Machines, as well as DBN-3 using AE for synthetic data generation, obtained similar results.



\begin{table*}[!htb]
\renewcommand{\arraystretch}{1.85}
\centering
\resizebox{\textwidth}{!}{
\begin{tabular}{llllllllllllllll}
\hline
& \multicolumn{3}{c}{\textbf{RBM}} & \multicolumn{3}{c}{\textbf{DBN-2}} & \multicolumn{3}{c}{\textbf{DBN-3}} \\ \hline
& Aug-RBM          & Aug-AE       & Baseline                  & Aug-RBM           & Aug-AE         & Baseline       & Aug-RBM   &    Aug-AE                &       Baseline     \\ \hline 
                          
\multicolumn{1}{c}{\textbf{ACC}}   & \textbf{87.51$\pm$0.14} & 75.85$\pm$0.13  & 86.21$\pm$0.30   & 86.97$\pm$0.31 & 87.01$\pm$0.22   & 85.97$\pm$0.50  & 85.97$\pm$0.59 & \textbf{87.29$\pm$0.37}  & 84.73$\pm$0.94 \\

\multicolumn{1}{c}{\textbf{BAC}}   & 77.84$\pm$0.82 & 43.85$\pm$0.84  & 63.77$\pm$1.15   & \textbf{78.84$\pm$1.22} & 73.83$\pm$0.74   & 62.97$\pm$2.88  & \textbf{77.66$\pm$1.88} & \textbf{77.87$\pm$1.58}  & 60.55$\pm$2.85 \\

\multicolumn{1}{c}{\textbf{Kappa}} & \textbf{0.736$\pm$0.004} & 0.368$\pm$0.009 & 0.662$\pm$0.006  & \textbf{0.731$\pm$0.007} & 0.710$\pm$0.005  & 0.659$\pm$0.012 & 0.711$\pm$0.010 & 0.724$\pm$0.009 & 0.615$\pm$0.023 \\ \hline

\end{tabular}}
\\~\\
\caption{Effectiveness over Protozoa dataset using the proposed approaches.}
\label{t.proto_summary}
\end{table*}

\subsection{Training Analysis}
\label{ss.trainingAnalysis}

\begin{figure}
	\centering
	\subfigure[]{\includegraphics[width=0.49\textwidth]{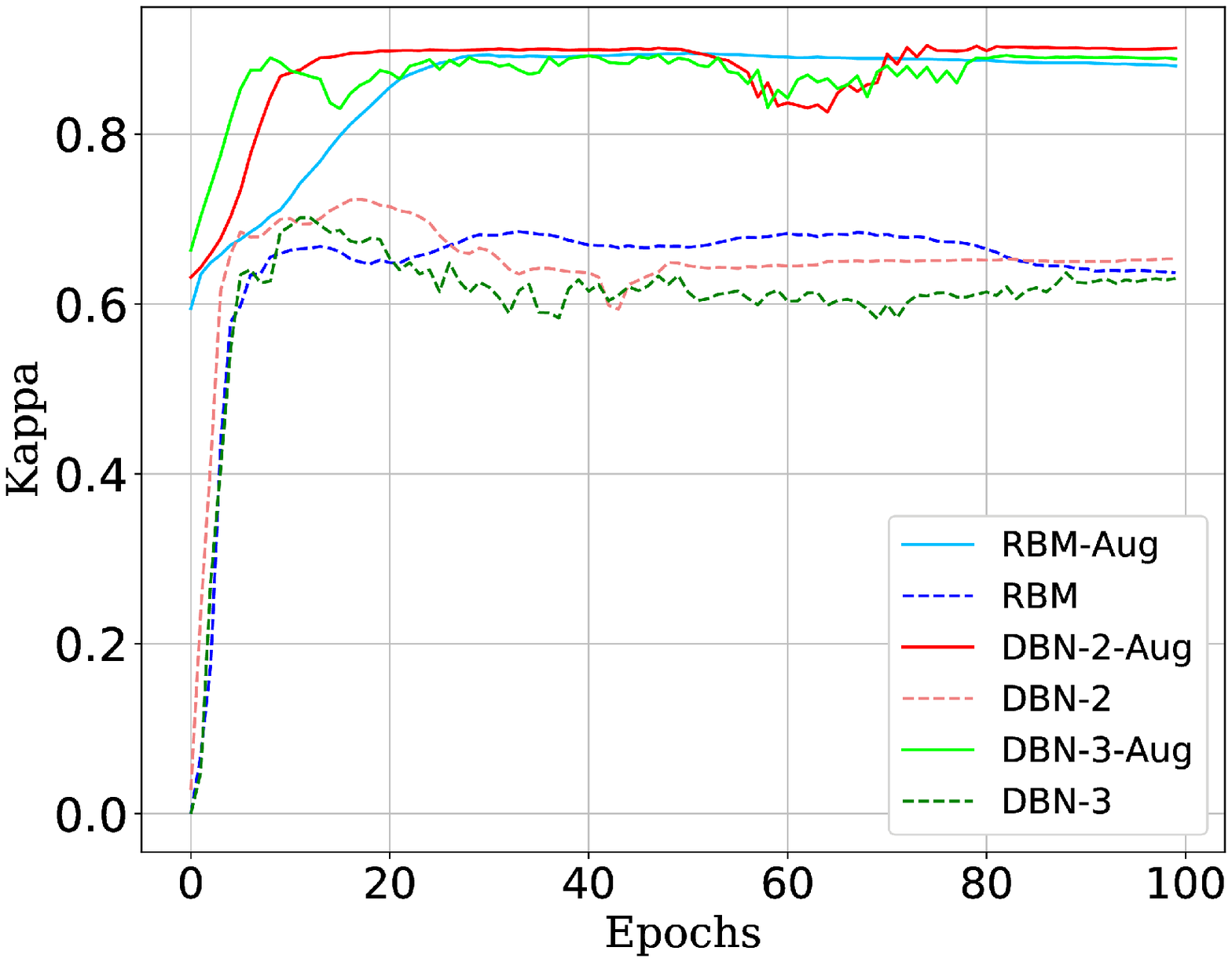}}
	\subfigure[]{\includegraphics[width=0.49\textwidth]{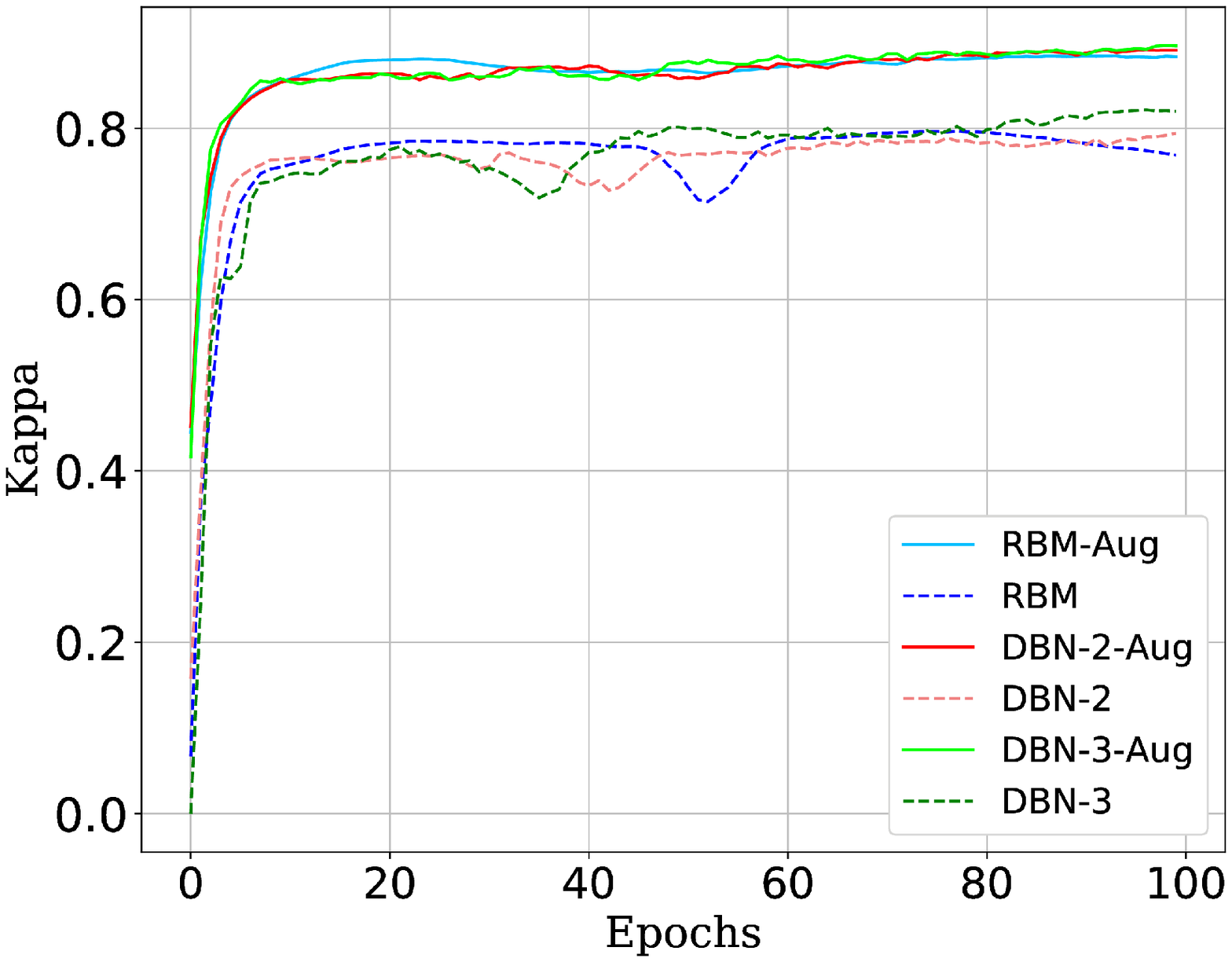}}
	\subfigure[]{\includegraphics[width=0.49\textwidth]{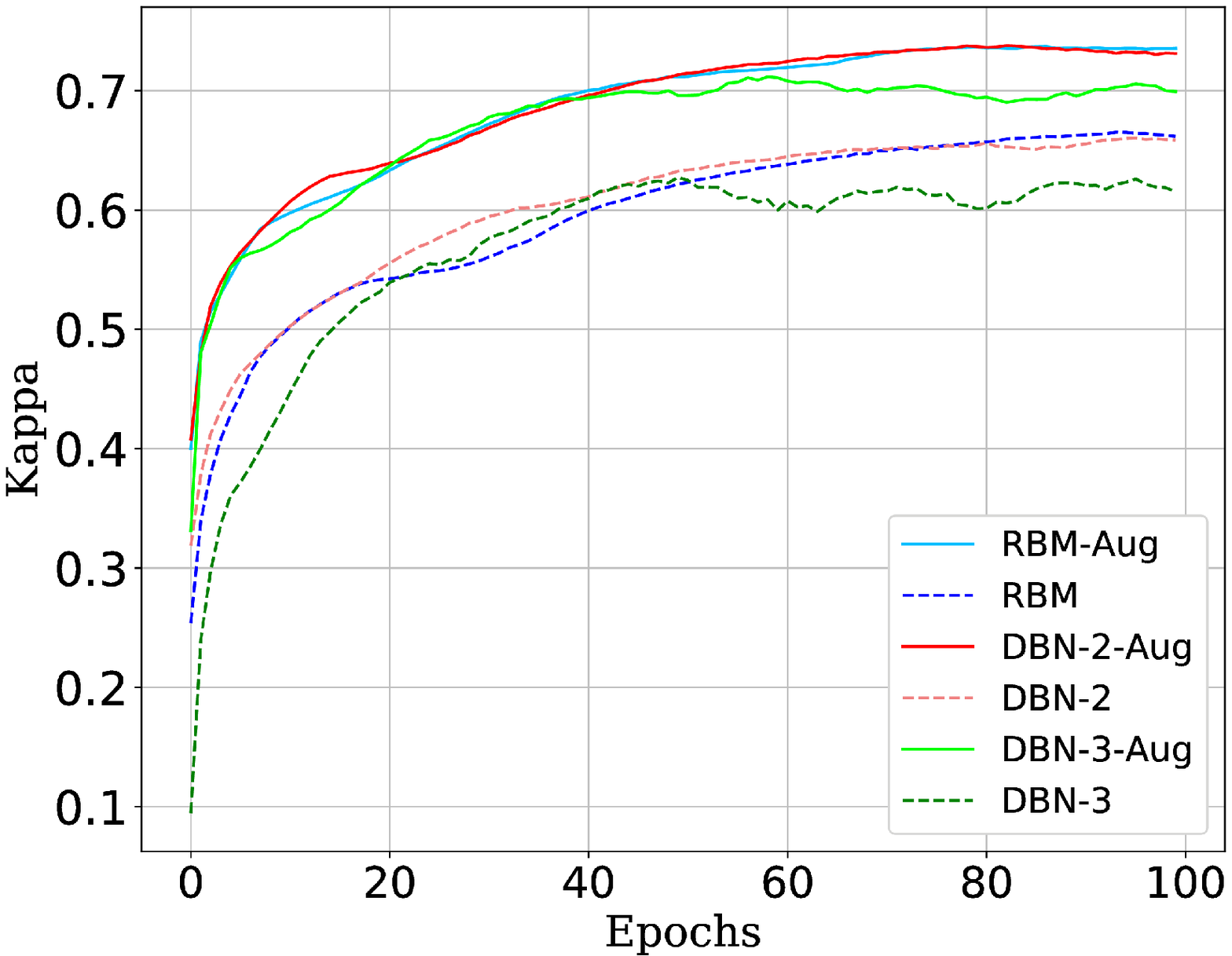}}

	\caption{Average Kappa values over the testing set concerning (a)~Larvae, (b)~Eggs, and (c)~Protozoa datasets.}
	\label{f.trainingProgress}
\end{figure}

Regarding the training analysis, we considered the datasets aumented with RBMs only since these models outperformed the ones using Autoencoders. Figure~\ref{f.trainingProgress} depicts the evolution of the Kappa values over the testing set during training. One can notice that: (i) data augmentation provided a considerable improvement in the results, (ii) training with data augmentation led to more stable results (Figures~\ref{f.trainingProgress}a and~\ref{f.trainingProgress}b), and (iii) differently from the other two datasets, techniques over Protozoa kept learning up to $80$ epochs (Figure~\ref{f.trainingProgress}c). Such behavior is somehow expected since Protozoa dataset poses a more challenging scenario. The stable results provided by data augmentation may allow us to apply some criteria for convergence analysis during training, such as early stop.
 
\subsection{Data Augmentation Analysis}
\label{ss.imageReconstruction}

\begin{figure}[!htb]
	\centerline{
		\begin{tabular}{cccccc}
			\includegraphics[width=1.9cm,height=1.9cm]{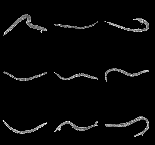} &
			\includegraphics[width=1.9cm,height=1.9cm]{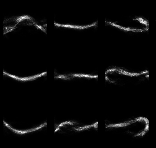} &
			\includegraphics[width=1.9cm,height=1.9cm]{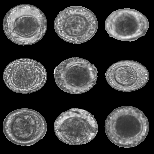} &
			\includegraphics[width=1.9cm,height=1.9cm]{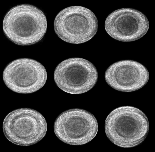} &
			\includegraphics[width=1.9cm,height=1.9cm]{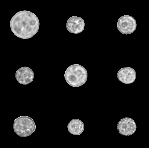} &
			\includegraphics[width=1.9cm,height=1.9cm]{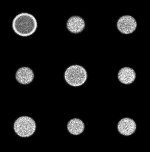} \\
			(a) & (b) & (c) & (d) & (e) & (f)
	\end{tabular}}
\caption{Data augmentation analysis: (a)~real and (b)~synthetic Larvae samples, (c)~real and (d)~synthetic Eggs samples, and (e)~real and (f)~synthetic Protozoa samples.}
\label{f.augmentation}
\end{figure}

Figure~\ref{f.augmentation} shows some synthetic data generated by RBMs using $500$ hidden neurons. One can observe that RBMs were able to generate useful samples, which corroborates the aforementioned results, i.e., such a process improved the parasites classification. Besides, the less accurate results concern the ones related to the Larvae dataset since we have a small subset of samples and their shape change considerably among the parasites.

\section{Conclusions and Future Works}
\label{s.conclusion}

This paper dealt with the problem of human intestinal parasites classification through RBM and DBN approaches. Experiments conducted over three distinct scenarios composed of Larvae, Eggs, and Protozoa, which are also partially surrounded by fecal impurities, confirmed the robustness of the models for classification purposes. Additionally, the performance of RBMs was also compared against Autoencoders for data augmentation since the datasets are highly unbalanced. Regarding future works, we intend to analyze the behavior of the models over a broader spectrum using colored images, as well as employing other RBM-based models, such as the Infinite RBMs (iRBMs) and the DBMs, to the task of human intestinal parasites classification.

\section*{Acknowledgments} 
The authors are grateful to FAPESP grants \#2013/07375-0, \#2014/12236-1, \#2017/25908-6, \#2019/07825-1, and \#2019/07665-4, as well as CNPq grants \#307066/2017-7, and \#427968/2018-6. This study was financed in part by the Coordena\c{c}\~ao de Aperfei\c{c}oamento de Pessoal de N\'ivel Superior – Brasil (CAPES) – Finance Code 001.

\bibliographystyle{splncs04}
\bibliography{references.bib}

\end{document}